\pdfoutput=1

\documentclass[11pt]{article}

\usepackage[]{emnlp2021}

\usepackage{times}
\usepackage{latexsym}

\usepackage{amsmath}
\usepackage{amssymb}
\usepackage{subcaption}
\usepackage{caption}
\usepackage{multicol}
\usepackage{multirow}
\usepackage{booktabs}
\usepackage{graphicx}
\usepackage{float}
\usepackage{efbox}
\usepackage{amssymb}
\usepackage{pifont}
\newcommand{\cmark}{\ding{51}}%
\newcommand{\xmark}{\ding{55}}%
\def\@fnsymbol#1{\ensuremath{\ifcase#1\or *\or \dagger\or \ddagger\or
   \mathsection\or \mathparagraph\or \|\or **\or \dagger\dagger
   \or \ddagger\ddagger \else\@ctrerr\fi}}
\newcommand{\ssymbol}[1]{^{\@fnsymbol{#1}}}

\DeclareMathOperator*{\argmin}{arg\,min}

\usepackage[T1]{fontenc}

\usepackage[utf8]{inputenc}

\usepackage{microtype}

\usepackage[ruled,vlined]{algorithm2e}

\title{Effectiveness of Pre-training for Few-shot Intent Classification}

\author{Haode Zhang$^1$\thanks{~ Equal contribution.} \quad Yuwei Zhang$^{1*}$  \quad Li-Ming Zhan$^{1}$ \\ 
\bf \quad Jiaxin Chen$^{1}$ \quad Guangyuan Shi$^{1}$  \quad Albert Y.S. Lam$^{2}$  \quad Xiao-Ming Wu$^{1}\Thanks{~ Corresponding author.}$\\
Department of Computing, The Hong Kong Polytechnic University, Hong Kong S.A.R.$^1$ \\
Fano Labs, Hong Kong S.A.R.$^2$ \\
{\tt haode.zhang@connect.polyu.hk},~ {\tt zhangyuwei.work@gmail.com} \\
{\tt \{lmzhan.zhan, jiax.chen, guang-yuan.shi\}@connect.polyu.hk} \\
{\tt xiao-ming.wu@polyu.edu.hk},~
{\tt albert@fano.ai}\\
}

\begin{document}
\maketitle
\begin{abstract}

This paper investigates the effectiveness of pre-training for few-shot intent classification. While existing paradigms commonly further pre-train language models such as BERT on a vast amount of unlabeled corpus, we find it highly effective and efficient to simply fine-tune BERT with a small set of labeled utterances from public datasets. Specifically, fine-tuning BERT with roughly 1,000 labeled data yields a pre-trained model -- IntentBERT, which can easily surpass the performance of existing pre-trained models for few-shot intent classification on novel domains with very different semantics. The high effectiveness of IntentBERT confirms the feasibility and practicality of few-shot intent detection, and its high generalization ability across different domains suggests that intent classification tasks may share a similar underlying structure, which can be efficiently learned from a small set of labeled data.
The source code can be found at \url{https://github.com/hdzhang-code/IntentBERT}.

\end{abstract}

\section{Introduction}
\efboxsetup{linecolor=black,linewidth=0.7pt}
\begin{figure}[t]
    \begin{subfigure}{.49\linewidth}
    \centering
        \efbox{\includegraphics[scale=0.165]{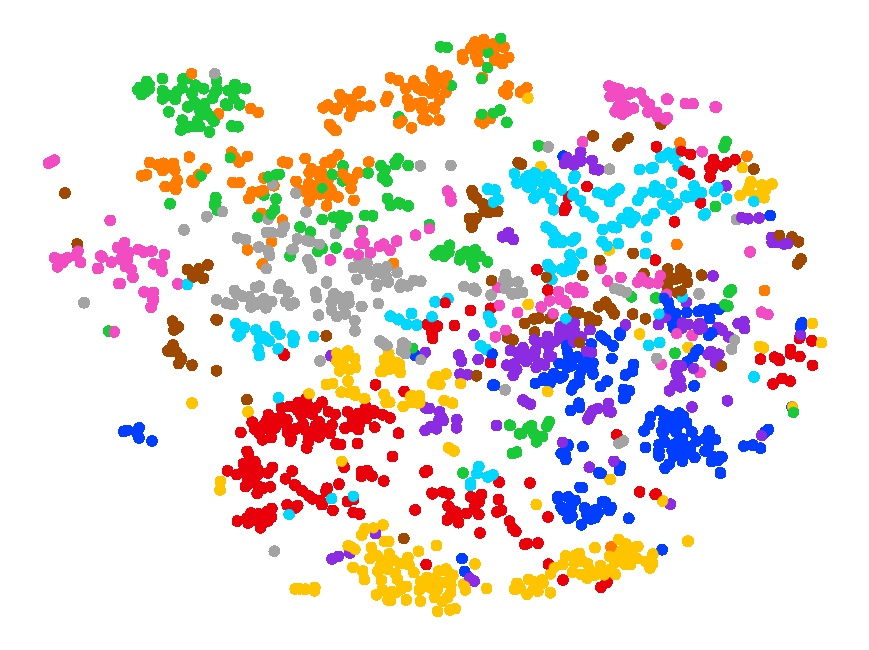}}
        \caption{BERT}
        \label{subfigure: scatter plot BERT}
    \end{subfigure}
    \begin{subfigure}{.49\linewidth}
    \centering
        \efbox{\includegraphics[scale=0.165]{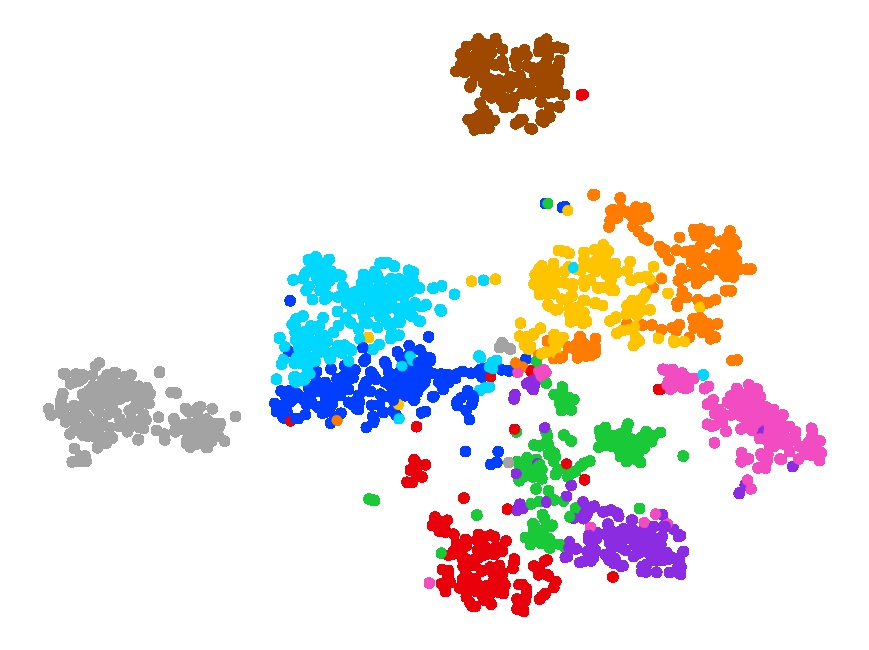}}
        \caption{TOD-BERT}
        \label{subfigure: scatter plot TOD-BERT}
    \end{subfigure}
    \newline
    \begin{subfigure}{.49\linewidth}
    \centering
        \efbox{\includegraphics[scale=0.165]{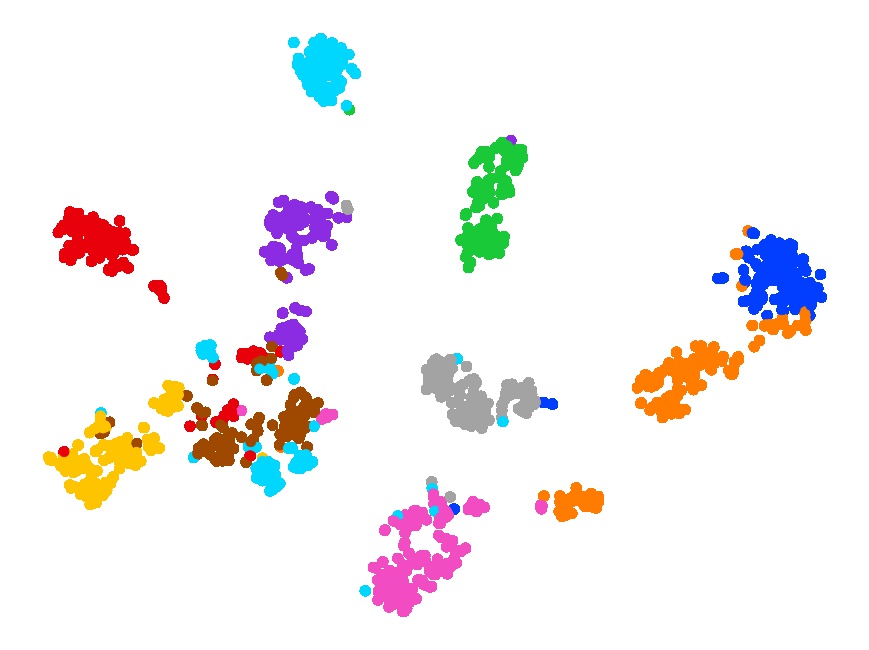}}
        \caption{IntentBERT (ours)}
        \label{subfigure: scatter plot IntentBERT}
    \end{subfigure}
    \begin{subfigure}{.49\linewidth}
    \centering
        \efbox{\includegraphics[scale=0.165]{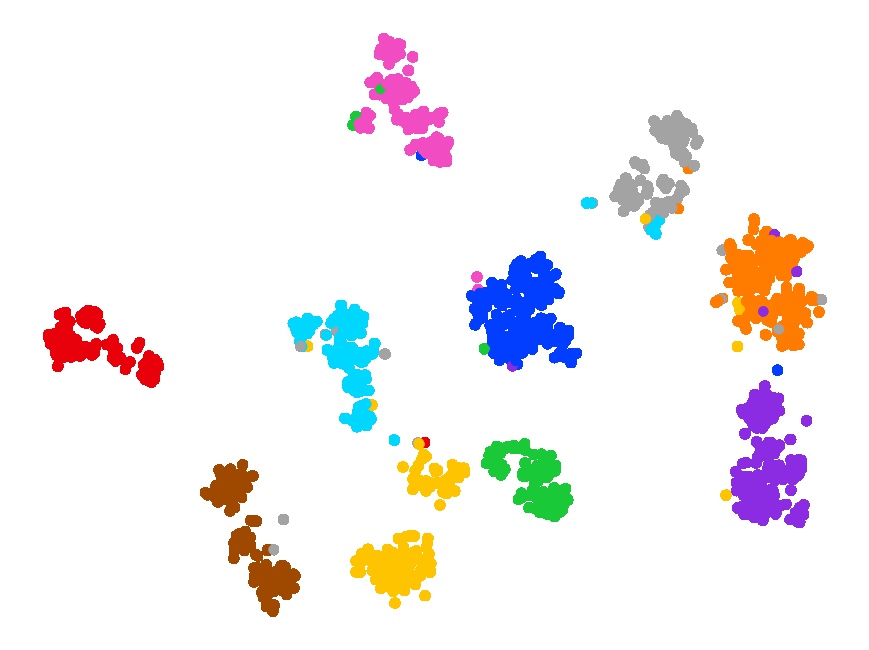}}
        \caption{IntentBERT+MLM (ours)}
        \label{subfigure: scatter plot IntentBERT+MLM}
    \end{subfigure}
    \caption{Visualization of the embedding spaces with t-SNE. We randomly sample 10 classes and 500 data per class from BANKING77 (best viewed in color).}
    \label{figure: scatter plot}
\end{figure}
Task-oriented dialogue systems have been widely deployed to a variety of sectors~\cite{yan2017building, chen2017survey, zhang2020recent, NEURIPS2020_e9462095}, ranging from shopping~\cite{yan2017building} to medical services~\cite{arora2020cross, wei-etal-2018-task}, to provide interactive experience. 
Training an accurate intent classifier is vital for the development of such task-oriented dialogue systems. However, an important issue is how to achieve this when only limited number of labeled instances are available, which is often the case at the early development stage.

To tackle few-shot intent detection, some recent attempts employ induction network~\citep{geng2019few}, generation-based methods~\citep{xia2020composed, xia-etal-2020-composed}, metric learning~\citep{nguyen2020dynamic}, or self-training~\citep{dopierre2020few}. 
These works mainly focus on designing novel algorithms for representation learning and inference, which often comes with complicated models. Most recently, large-scale pre-trained language models such as BERT~\citep{devlin2018bert, radford2019language, brown2020language} have shown great promise in many natural language understanding tasks~\cite{DBLP:conf/nips/WangPNSMHLB19}, and there has been a surge of interest in fine-tuning the pre-trained language models for intent detection~\citep{zhang2020discriminative, zhang2020intent, peng2020soloist, wu2020tod, casanueva2020efficient,larson2019evaluation}.

While fine-tuning pre-trained language models on large-scale annotated datasets has yielded significant improvements in many tasks including intent detection, it is laborious and expensive to construct large-scale annotated datasets in new application domains.
Therefore, recent efforts have been dedicated to adapting pre-trained language models to a specific task such as intent detection by conducting continued pre-training~\cite{gururangan2020don, gu-etal-2021-pral} on a large unlabeled dialogue corpus with a specially designed optimization objective. Below we summarize the most related works in this line of research for few-shot intent detection.

\begin{itemize}
    \item \textbf{CONVBERT}~\cite{mehri2020dialoglue} finetunes BERT on an unlabeled dialogue corpus consisting of nearly $700$ million conversations.
    \item \textbf{TOD-BERT}~\cite{wu2020tod} further pre-trains BERT on a task-oriented dialogue corpus of $100,000$ unlabeled samples with masked language modelling~(MLM) and response contrastive objectives.
    \item \textbf{USE-ConveRT}~\cite{henderson-etal-2020-convert, casanueva2020efficient} investigates a dual encoder model trained with response selection tasks on $727$ million input-response pairs.
    \item \textbf{DNNC}~\cite{zhang2020discriminative} pre-trains a language model with around $1$ million annotated samples for natural language inference~(NLI) and use the pre-trained model for intent detection. 
    \item \textbf{WikiHowRoBERTa}~\cite{zhang2020intent} constructs some pre-training tasks based on the wikiHow database with $110,000$ articles.
\end{itemize}

While these methods have achieved impressive performance, they heavily rely on the existence of a large-scale corpus~\cite{mehri2020dialoglue} that is close in semantics to the target domain or consists of similar tasks
for continued pre-training, which needs huge effort for data collection and comes at a high computational cost. More importantly, they completely ignore the ``free lunch'' --
the publicly available, high-quality, manually-annotated intent detection benchmarks. For example, the dataset OOS~\cite{larson2019evaluation} provides labeled utterances across $10$ different domains. 
Hence, our study in this paper centers around the following research question: 
\begin{itemize}
    \item Is it possible to utilize publicly available datasets to pre-train an intent detection model that can  \emph{learn transferable task-specific knowledge to generalize across different domains}?
\end{itemize}

In this paper, we provide an affirmative answer to this question. We fine-tune BERT using a simple standard supervised training with approximately 1,000 labeled utterances from public datasets and obtain a pre-trained model, called IntentBERT. It can be directly applied for few-shot intent classification on a target domain that is drastically different from the pre-training data and 
significantly outperform existing pre-trained models,
without further fine-tuning on target data (labeled or unlabeled). This simple ``free-lunch'' solution not only confirms the feasibility and practicality of few-shot intent detection, but also provides a ready-to-use well-performing model for practical use, saving the effort in algorithm design and data collection. Moreover, the high generalization ability of IntentBERT on cross-domain few-shot classification tasks, which are generally considered very difficult due to large domain gaps and the few data constraint, suggests that most intent detection tasks probably share a common underlying structure that could be learned from a small set of data.

Further, to leverage unlabeled data in the target domain, we design a joint pre-training scheme, which simultaneously optimizes the classification error on the source labeled data and the language modeling loss on the target unlabeled data. This joint-training scheme can learn better semantic representations and significantly outperforms existing two-stage pre-training methods~\cite{gururangan2020don}. A visualization of the embedding spaces produced by strong baselines and our methods is provided in Fig.~\ref{figure: scatter plot}, which clearly demonstrates the superiority of our pre-trained models.

\section{Methodology}
We present a continued pre-training framework for intent classification based on the pre-trained language model BERT~\cite{devlin2018bert}.

Our pre-training method relies on the existence of a small labeled dataset $\mathcal{D}_{\text{source}}^{\text{labeled}} = \{(x_i, y_i)\}$, where $y_i$ is the label of utterance $x_i$. Such data samples can be readily obtained from public intent detection datasets such as OOS~\cite{larson2019evaluation} and HWU64~\cite{liu2021benchmarking}. As will be shown in the experiments, roughly $1,000$ examples from either OOS or HWU64 are enough for the pre-trained intent detection model to achieve a superior performance on drastically different target domains such as ``Covid-19''.

We further consider a scenario that unlabeled utterances $\mathcal{D}_{\text{target}}^{\text{unlabeled}} = \{x_i\}$ in the target domain are available, and propose a joint pre-training scheme that is empirically proven to be highly effective.

\subsection{Supervised Pre-training}
 
Given $\mathcal{D}_{\text{source}}^{\text{labeled}}=\{(x_i,y_i)\}$ with $N$ different classes, we employ a simple method to fine-tune BERT. Specifically, a linear layer is attached on top of BERT as the classifier, i.e., 
\begin{equation}
    \begin{split}
        p(y|h_i) = \text{softmax}\left(\mathbf{W}h_i + \mathbf{b}\right) \in \mathbb{R}^N,
    \end{split}
    \label{equation: transfer learning task}
\end{equation}
where $h_i \in \mathbb{R}^{d}$ is the feature representation of $x_i$ given by the $[CLS]$ token, $\mathbf{W} \in \mathbb{R}^{N \times d}$ and $\mathbf{b} \in \mathbb{R}^{N}$ are parameters of the linear layer. The model parameters $\theta=\left\{\phi, \mathbf{W}, \mathbf{b}\right\}$, with $\phi$ being the parameters of BERT, are trained on $\mathcal{D}_{\text{source}}^{\text{labeled}}$ with a cross-entropy loss:
\begin{equation}
    \setlength{\abovedisplayskip}{10pt}
    \setlength{\belowdisplayskip}{10pt}
    \begin{split}
        \theta^{*} = \argmin_{\theta} \mathcal{L}_{ce}\left(\mathcal{D}_{\text{source}}^{\text{labeled}};\theta\right).
    \end{split}
    \label{equation: transfer learning task}
\end{equation}

After training, the fine-tuned BERT is expected to have learned general intent detection skills, and hence we call it IntentBERT.

\subsection{Joint Pre-training}
\label{methodology: joint loss, MLM}
Given unlabeled target data $\mathcal{D}_{\text{target}}^{\text{unlabeled}}$, we can leverage it to further enhance our IntentBERT, by simultaneously optimizing a language modeling loss on $\mathcal{D}_{\text{target}}^{\text{unlabeled}}$ and the supervised loss in Eq.~(\ref{equation: transfer learning task}). The language modeling loss can help to learn semantic representations of the target domain while preventing overfitting to the source data.

Specifically, we use MLM as the language modeling loss, in which a proportion of input tokens are masked with the special token $[MASK]$ and the model is trained to retrieve the masked tokens. The joint training loss is formulated as:
\begin{equation}
    \mathcal{L}_{\text{joint}}=\mathcal{L}_{ce}(\mathcal{D}_{\text{source}}^{\text{labeled}};\theta)+\lambda \mathcal{L}_{mlm}(\mathcal{D}_{\text{target}}^{\text{unlabeled}};\theta),
    \label{equation: final loss}
\end{equation}
where $\lambda$ is a hyperparameter that balances the supervised loss and the unsupervised loss.

\subsection{Few-shot Intent Classification}
After pre-training, the parameters of IntentBERT are fixed, and it can be immediately used as a feature extractor for novel few-shot intent classification tasks. The classifier can be a parametric one such as logistic regression or a non-parametric one such as nearest neighbor. A parametric classifier will be trained with the few labeled examples provided in a task and make predictions on the unlabeled queries. As will be shown in the experiments, a simple linear classifier suffices to achieve very good performance, thanks to the effective utterance representations produced by IntentBERT. 

    \begin{table*}[t]
\centering
\small
\begin{tabular}{lccccccc}
\toprule
\multicolumn{1}{l}{\multirow{2}{*}{Method}} & 
\multicolumn{1}{c}{\multirow{2}{*}{$\mathcal{D}_{\text{target}}^{\text{unlabeled}}$}} & 
\multicolumn{2}{c}{BANKING77}  &
\multicolumn{2}{c}{MCID}    &
\multicolumn{2}{c}{HINT3}  
\\
\cmidrule(lr){3-4}  \cmidrule(lr){5-6}  \cmidrule(lr){7-8}
& & 2-shot  & 10-shot  & 2-shot & 10-shot & 2-shot & 10-shot  \\
\midrule
BERT-Freeze & \xmark & 52.6\tiny{$~\pm$12.4} & 70.0\tiny{$~\pm$11.7} & 57.8\tiny{$~\pm$11.7} & 72.4\tiny{$~\pm$10.7} & 47.3\tiny{$~\pm$12.1} & 66.8\tiny{$~\pm$10.5} \\
CONVBERT & \xmark & 68.3\tiny{$~\pm$12.3} & 86.6\tiny{$~\pm$8.2} & 67.7\tiny{$~\pm$11.5} & 83.5\tiny{$~\pm$7.9} & 72.6\tiny{$~\pm$10.9} & 87.2\tiny{$~\pm$7.9} \\
TOD-BERT & \xmark & 77.7\tiny{$~\pm$7.4} & 89.4 \tiny{$~\pm$5.1} &  64.1\tiny{$~\pm$9.0}  & 77.7\tiny{$~\pm$11.0} &  68.9\tiny{$~\pm$11.7} &  83.5\tiny{$~\pm$8.6}  \\
USE-ConveRT$\ssymbol{5}$ & \xmark  & -- & 85.2 & -- & -- & -- & -- \\
DNNC & \xmark   & 67.5\tiny{$~\pm$15.4} & 89.8\tiny{$~\pm$7.5} & 56.2\tiny{$~\pm$16.7} & 80.0\tiny{$~\pm$9.9} & 64.1\tiny{$~\pm$14.8} & 87.9\tiny{$~\pm$8.1}
\\
WikiHowRoBERTa & \xmark  & 34.9\tiny{$~\pm$10.5} & 41.6\tiny{$~\pm$10.1} & 30.8\tiny{$~\pm$9.9} & 36.4\tiny{$~\pm$9.7} & 31.7\tiny{$~\pm$10.3} & 39.0\tiny{$~\pm$9.9}
\\
\midrule
IntentBERT (HWU64) (ours) & \xmark  & \textbf{78.4\tiny{$~\pm$10.6}}  & \textbf{90.0\tiny{$~\pm$7.5}}  & \textbf{74.5\tiny{$~\pm$11.9}}  & \textbf{85.9\tiny{$~\pm$8.8}}  & \textbf{77.9\tiny{$~\pm$10.6}}  & \textbf{89.4\tiny{$~\pm$7.9}} \\
IntentBERT (OOS) (ours) & \xmark  & \textbf{82.4\tiny{$~\pm$8.3}} & \textbf{91.8\tiny{$~\pm$4.2}}  & \textbf{77.1\tiny{$~\pm$9.0}} & \textbf{88.1\tiny{$~\pm$5.9}}  & \textbf{80.1\tiny{$~\pm$10.4}} & \textbf{90.2\tiny{$~\pm$7.4}}  \\
IntentBERT (OOS)+MLM (ours) & \cmark  & \textbf{88.9\tiny{$~\pm$9.0}} & \textbf{95.2\tiny{$~\pm$5.1}}  & \textbf{86.3\tiny{$~\pm$9.8}} & \textbf{92.4\tiny{$~\pm$6.2}}  &  \textbf{87.1\tiny{$~\pm$9.8}} & \textbf{94.0\tiny{$~\pm$6.0}} \\
\bottomrule
\end{tabular}
\caption{Main results for $5$-way tasks. $\ssymbol{5}$ stands for results from the original paper.}
\label{table: main result, effectiveness of our methods}
\end{table*}
\section{Experiments}
\subsection{Experimental Setup}
\begin{figure}[H]
     \centering
     \includegraphics[scale = 0.35]{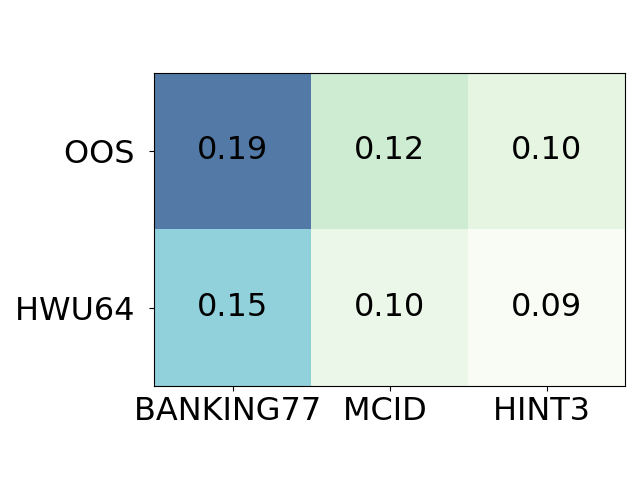}
\caption{Vocabulary overlap.}
\label{figure: Domain relevance measured by vocabulary overlap.}
\end{figure}
\begin{table}[h]
\centering
\small
\begin{tabular}{lccc}
\toprule
Dataset & \#domain & \#intent & \#utterances \\
\midrule
OOS          & 8     & 120   & 18000 \\
HWU64        & 21     & 64    & 25716    \\  
\midrule
BANKING77   & 1   & 77   & 13083 \\
MCID        & 1     & 16   & 1745    \\  
HINT3       & 3    & 51   & 2011    \\  
\bottomrule
\end{tabular}
\caption{Dataset statistics.}
\label{table: Dataset statistics}
\end{table}
\textbf{Datasets}. To train our IntentBERT, we continue to pre-train BERT on either of the two datasets, OOS~\cite{larson2019evaluation}\footnote{The domains ``Banking'' and ``Credit Cards'' are excluded because they are semantically close to the evaluation data.} and HWU64~\cite{liu2021benchmarking}, both of which contain multiple domains, providing rich resources to learn from\footnote{We have also experimented with the combination of both datasets but observed no better results.}. For evaluation, we employ three datasets: 
\textbf{BANKING77}~\cite{casanueva2020efficient} is a fine-grained intent detection dataset focusing on ``Banking''; \textbf{MCID} ~\cite{arora2020cross} is a dataset for ``Covid-19'' chat bots; \textbf{HINT3}~\cite{arora-etal-2020-hint3} contains $3$ domains, ``Mattress Products Retail'', ``Fitness Supplements Retail'' and ``Online Gaming''. Dataset statistics are summarized in Table~\ref{table: Dataset statistics}.

Fig.~\ref{figure: Domain relevance measured by vocabulary overlap.} visualizes the vocabulary overlap between the source training data and target test data, which is calculated as the proportion of the shared words in the combined vocabulary of any two datasets after removing stop words. It is observed that the overlaps are quite small, 
indicating the existence of large semantic gaps.

\textbf{Evaluation}. The classification performance is evaluated by $C$-way $K$-shot tasks. For each task, We randomly sample $C$ classes and $K$ examples per class to train the classifier, and then we sample extra $5$ examples per class as queries for evaluation. The accuracy is averaged over $500$ such tasks.

\textbf{Baselines.} We compare IntentBERT to the following strong baselines. \textbf{BERT-Freeze} simply freeze the off-the-shelf BERT; 
\textbf{TOD-BERT}~\cite{wu2020tod} further pre-trains BERT on a huge amount of task-oriented conversations with MLM and response selection tasks; \textbf{CONVBERT}~\cite{mehri2020dialoglue} further pre-trains BERT on a large open-domain multi-turn dialogue corpus; \textbf{USE-ConveRT}~\cite{henderson-etal-2020-convert, casanueva2020efficient} is a fast embedding-based classifier pre-trained on an open-domain dialogue corpus by dialogue response selection tasks; \textbf{DNNC}~\cite{zhang2020discriminative} further pre-trains a BERT-based model on NLI tasks and then applies a similarity-based classifier for classification;
\textbf{WikiHowRoBERTa}~\cite{zhang2020intent} further pre-trains RoBERTa~\cite{liu2019roberta} on fake intent detection data synthesized from wikiHow\footnote{https://www.wikihow.com/}.

All the baselines (except BERT-Freeze) adopt a second pre-training stage, but with different objectives and on different corpus. In our experiments, all the baselines (except DNNC) use logistic regression as the classifier. For DNNC, we strictly follow the original implementation\footnote{https://github.com/salesforce/DNNC-few-shot-intent} to 
pre-train a BERT-style pairwise encoder to estimate the best matched training example for a query utterance.

\textbf{Training details.}
We use BERT\textsubscript{base}\footnote{https://github.com/huggingface/transformers}~(the base configuration with
$d=768$) as the encoder, Adam \cite{kingma2014adam} as the optimizer, and PyTorch library for implementation. The model is trained with Nvidia GeForce RTX 2080 Ti GPUs. For supervised pre-training, we use validation to control early-stop to prevent overfitting. Specifically, we use HWU64 for validation when pre-training with OOS and vice versa. The training is stopped if no improvement in accuracy is observed in $3$ epochs. For joint pre-training, $\lambda$ is set to $1$. The number of training epochs is fixed to $10$, since it is not prone to overfitting.

\subsection{Main Results}
The main results are provided in Table~\ref{table: main result, effectiveness of our methods}. First, IntentBERT (either pre-trained with OOS or HWU64) consistently outperforms all the baselines by a significant margin in most cases. Take the results of $5$-way $2$-shot classification on MCID for example, IntentBERT (OOS) outperforms the strongest baseline CONVBERT by an absolute margin of $9.4\%$, demonstrating the high effectiveness of our pre-training method. The cross-domain transferability of IntentBERT indicates that despite semantic domain gaps, most intent detection tasks probably share a similar underlying structure, which could be learned with a small set of labeled utterances. 
Second, IntentBERT (OOS) seems to be more effective than IntentBERT (HWU64), which may be due to the semantic diversity of the training corpus. Nevertheless, the small difference in performance between them shows that our pre-training method is not sensitive to the training corpus.
\begin{figure}[h]
     \centering
     \includegraphics[scale = 0.40]{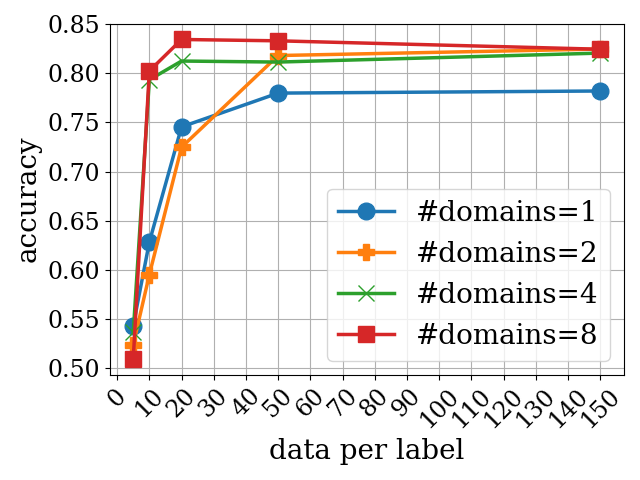}
\caption{Effect of the amount of labeled data used for pre-training in the source domain (OOS). The results are evaluated on $5$-way $2$-shot tasks on BANKING77.}
\label{figure: impact of training data on the transferring performance of IntentBERT}
\end{figure}

Finally, our proposed joint pre-training scheme (Section~\ref{methodology: joint loss, MLM}) achieves significant improvement over IntentBERT (up to 9.2\% absolute margin), showing the high effectiveness of joint pre-training when target unlabeled data is accessible. Our joint pre-training scheme can also be applied to other language models such as GPT-2~\cite{radford2019language} and ELMo~\cite{peters-etal-2018-deep}, which is left as future work.

\subsection{Analysis}
\label{section: How Much Data Is Needed}

\textbf{Amount of labeled data for pre-training}. We reduce the data used for pre-training in two dimensions: the number of domains and the number of samples per class. We randomly sample $1$, $2$, $4$ and $8$ domains for multiple times and report the averaged results in Fig.~\ref{figure: impact of training data on the transferring performance of IntentBERT}. It is found that the training data can be dramatically reduced without harming the performance. 
The model trained on $4$ domains and $20$ samples per class performs on par with that on $8$ domains and $150$ samples per class. In general, we only need around $1,000$ annotated utterances to train IntentBERT, which can be easily obtained in public datasets.
This finding indicates that using small task-relevant data for pre-training may be a more effective and efficient fine-tuning paradigm.

\begin{figure}[h]
     \centering
     \includegraphics[scale = 0.45]{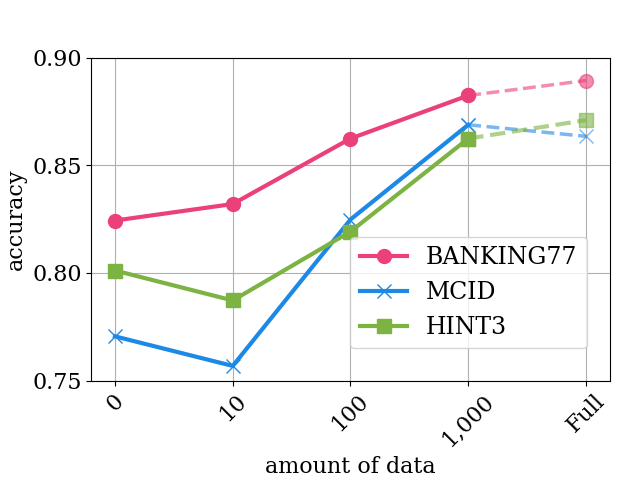}
\caption{Effect of the amount of unlabeled data used for joint pre-training in the target domain. The results are evaluated on $5$-way $2$-shot tasks with OOS as the source dataset.}
\label{figure: mlm data amount}
\end{figure}

\textbf{Amount of unlabeled data for joint pre-training}. We randomly sample a fraction of unlabeled utterances and re-run the joint training. As shown in Fig.~\ref{figure: mlm data amount}, the accuracy keeps increasing when the number of unlabeled samples grows from $10$ to $1,000$ and tends to saturate after reaching $1,000$. Surprisingly, $1,000$ utterances in BANKING77 can yield a comparable performance than the full dataset ($13,083$ utterances). Generally, it does not need much unlabeled data to reach a high accuracy.

\textbf{Ablation study on joint pre-training.} First, we investigate a two-stage pre-training scheme~\cite{gururangan2020don} where we use BERT or IntentBERT as initialization and perform MLM in the target domain (the top two rows in Table~\ref{table: Ablation study}). It can be seen that they perform much worse than our joint pre-training scheme (the bottom row).
Second, we use the source data instead of the target data for MLM in joint pre-training (the third row),
and observe consistent performance drops, which shows the necessity of a domain-specific corpus. 
\begin{table}[h]
\centering
\small
\begin{tabular}{p{0.20\textwidth}p{0.04\textwidth}p{0.04\textwidth}p{0.04\textwidth}}
\toprule
Methods & BANK & MCID & HINT3\\
\midrule
BERT$\rightarrow$MLM(target) & 80.5 & 63.0 & 72.0\\
IntentBERT$\rightarrow$MLM(target) & 82.0 & 75.9 & 77.9\\
IntentBERT+MLM(source) & 84.1 & 75.9 & 78.5 \\
IntentBERT+MLM(target) & \textbf{88.9} & \textbf{86.3} & \textbf{87.1} \\
\bottomrule
\end{tabular}
\caption{Ablation study on joint pre-training. BANK denotes BANKING77. $\rightarrow$ denotes moving to the next training stage. $+$ denotes joint optimization of both loss functions. The data used for the experiment (either from "target" or "source") is shown in the brackets. The results are evaluated on 5-way 2-shot tasks with OOS as the source dataset.}
\label{table: Ablation study}
\end{table}


\section{Conclusion}

We have proposed IntentBERT, a pre-trained model for few-shot intent classification, which is obtained by fine-tuning BERT on a small set of publicly available labeled utterances. We have shown that using small task-relevant data for fine-tuning is far more effective and efficient than current practice that fine-tunes on a large labeled or unlabeled dialogue corpus. This finding may have a wide implication for other tasks besides intent detection. 

\section*{Acknowledgments}
We would like to thank the anonymous reviewers for their helpful comments. This research was supported by the grants of HK ITF UIM/377 and DaSAIL project P0030935.


\begin{thebibliography}{30}
\expandafter\ifx\csname natexlab\endcsname\relax\def\natexlab#1{#1}\fi

\bibitem[{Arora et~al.(2020{\natexlab{a}})Arora, Shrivastava, Mohit, Lecanda,
  and Aly}]{arora2020cross}
Abhinav Arora, Akshat Shrivastava, Mrinal Mohit, Lorena Sainz-Maza Lecanda, and
  Ahmed Aly. 2020{\natexlab{a}}.
\newblock Cross-lingual transfer learning for intent detection of covid-19
  utterances.

\bibitem[{Arora et~al.(2020{\natexlab{b}})Arora, Jain, Chaturvedi, and
  Modi}]{arora-etal-2020-hint3}
Gaurav Arora, Chirag Jain, Manas Chaturvedi, and Krupal Modi.
  2020{\natexlab{b}}.
\newblock \href {https://doi.org/10.18653/v1/2020.insights-1.16} {{HINT}3:
  Raising the bar for intent detection in the wild}.
\newblock In \emph{Proceedings of the First Workshop on Insights from Negative
  Results in NLP}, pages 100--105, Online. Association for Computational
  Linguistics.

\bibitem[{Brown et~al.(2020)Brown, Mann, Ryder, Subbiah, Kaplan, Dhariwal,
  Neelakantan, Shyam, Sastry, Askell, Agarwal, Herbert{-}Voss, Krueger,
  Henighan, Child, Ramesh, Ziegler, Wu, Winter, Hesse, Chen, Sigler, Litwin,
  Gray, Chess, Clark, Berner, McCandlish, Radford, Sutskever, and
  Amodei}]{brown2020language}
Tom~B. Brown, Benjamin Mann, Nick Ryder, Melanie Subbiah, Jared Kaplan,
  Prafulla Dhariwal, Arvind Neelakantan, Pranav Shyam, Girish Sastry, Amanda
  Askell, Sandhini Agarwal, Ariel Herbert{-}Voss, Gretchen Krueger, Tom
  Henighan, Rewon Child, Aditya Ramesh, Daniel~M. Ziegler, Jeffrey Wu, Clemens
  Winter, Christopher Hesse, Mark Chen, Eric Sigler, Mateusz Litwin, Scott
  Gray, Benjamin Chess, Jack Clark, Christopher Berner, Sam McCandlish, Alec
  Radford, Ilya Sutskever, and Dario Amodei. 2020.
\newblock \href
  {https://proceedings.neurips.cc/paper/2020/hash/1457c0d6bfcb4967418bfb8ac142f64a-Abstract.html}
  {Language models are few-shot learners}.
\newblock In \emph{Advances in Neural Information Processing Systems 33: Annual
  Conference on Neural Information Processing Systems 2020, NeurIPS 2020,
  December 6-12, 2020, virtual}.

\bibitem[{Casanueva et~al.(2020)Casanueva, Tem{\v{c}}inas, Gerz, Henderson, and
  Vuli{\'c}}]{casanueva2020efficient}
I{\~n}igo Casanueva, Tadas Tem{\v{c}}inas, Daniela Gerz, Matthew Henderson, and
  Ivan Vuli{\'c}. 2020.
\newblock \href {https://doi.org/10.18653/v1/2020.nlp4convai-1.5} {Efficient
  intent detection with dual sentence encoders}.
\newblock In \emph{Proceedings of the 2nd Workshop on Natural Language
  Processing for Conversational AI}, pages 38--45, Online. Association for
  Computational Linguistics.

\bibitem[{Chen et~al.(2017)Chen, Liu, Yin, and Tang}]{chen2017survey}
Hongshen Chen, Xiaorui Liu, Dawei Yin, and Jiliang Tang. 2017.
\newblock A survey on dialogue systems: Recent advances and new frontiers.
\newblock \emph{Acm Sigkdd Explorations Newsletter}, 19(2):25--35.

\bibitem[{Devlin et~al.(2019)Devlin, Chang, Lee, and
  Toutanova}]{devlin2018bert}
Jacob Devlin, Ming-Wei Chang, Kenton Lee, and Kristina Toutanova. 2019.
\newblock \href {https://doi.org/10.18653/v1/N19-1423} {{BERT}: Pre-training of
  deep bidirectional transformers for language understanding}.
\newblock In \emph{Proceedings of the 2019 Conference of the North {A}merican
  Chapter of the Association for Computational Linguistics: Human Language
  Technologies, Volume 1 (Long and Short Papers)}, pages 4171--4186,
  Minneapolis, Minnesota. Association for Computational Linguistics.

\bibitem[{Dopierre et~al.(2020)Dopierre, Gravier, Subercaze, and
  Logerais}]{dopierre2020few}
Thomas Dopierre, Christophe Gravier, Julien Subercaze, and Wilfried Logerais.
  2020.
\newblock \href {https://doi.org/10.18653/v1/2020.coling-main.438} {Few-shot
  pseudo-labeling for intent detection}.
\newblock In \emph{Proceedings of the 28th International Conference on
  Computational Linguistics}, pages 4993--5003, Barcelona, Spain (Online).
  International Committee on Computational Linguistics.

\bibitem[{Geng et~al.(2019)Geng, Li, Li, Zhu, Jian, and Sun}]{geng2019few}
Ruiying Geng, Binhua Li, Yongbin Li, Xiaodan Zhu, Ping Jian, and Jian Sun.
  2019.
\newblock \href {https://doi.org/10.18653/v1/D19-1403} {Induction networks for
  few-shot text classification}.
\newblock In \emph{Proceedings of the 2019 Conference on Empirical Methods in
  Natural Language Processing and the 9th International Joint Conference on
  Natural Language Processing (EMNLP-IJCNLP)}, pages 3904--3913, Hong Kong,
  China. Association for Computational Linguistics.

\bibitem[{Gu et~al.(2021)Gu, Wu, Wu, Shi, and Yu}]{gu-etal-2021-pral}
Jing Gu, Qingyang Wu, Chongruo Wu, Weiyan Shi, and Zhou Yu. 2021.
\newblock \href {https://doi.org/10.18653/v1/2021.acl-short.40} {{PRAL}: A
  tailored pre-training model for task-oriented dialog generation}.
\newblock In \emph{Proceedings of the 59th Annual Meeting of the Association
  for Computational Linguistics and the 11th International Joint Conference on
  Natural Language Processing (Volume 2: Short Papers)}, pages 305--313,
  Online. Association for Computational Linguistics.

\bibitem[{Gururangan et~al.(2020)Gururangan, Marasovi{\'c}, Swayamdipta, Lo,
  Beltagy, Downey, and Smith}]{gururangan2020don}
Suchin Gururangan, Ana Marasovi{\'c}, Swabha Swayamdipta, Kyle Lo, Iz~Beltagy,
  Doug Downey, and Noah~A. Smith. 2020.
\newblock \href {https://doi.org/10.18653/v1/2020.acl-main.740} {Don{'}t stop
  pretraining: Adapt language models to domains and tasks}.
\newblock In \emph{Proceedings of the 58th Annual Meeting of the Association
  for Computational Linguistics}, pages 8342--8360, Online. Association for
  Computational Linguistics.

\bibitem[{Henderson et~al.(2020)Henderson, Casanueva, Mrk{\v{s}}i{\'c}, Su,
  Wen, and Vuli{\'c}}]{henderson-etal-2020-convert}
Matthew Henderson, I{\~n}igo Casanueva, Nikola Mrk{\v{s}}i{\'c}, Pei-Hao Su,
  Tsung-Hsien Wen, and Ivan Vuli{\'c}. 2020.
\newblock \href {https://doi.org/10.18653/v1/2020.findings-emnlp.196}
  {{C}onve{RT}: Efficient and accurate conversational representations from
  transformers}.
\newblock In \emph{Findings of the Association for Computational Linguistics:
  EMNLP 2020}, pages 2161--2174, Online. Association for Computational
  Linguistics.

\bibitem[{Hosseini{-}Asl et~al.(2020)Hosseini{-}Asl, McCann, Wu, Yavuz, and
  Socher}]{NEURIPS2020_e9462095}
Ehsan Hosseini{-}Asl, Bryan McCann, Chien{-}Sheng Wu, Semih Yavuz, and Richard
  Socher. 2020.
\newblock \href
  {https://proceedings.neurips.cc/paper/2020/hash/e946209592563be0f01c844ab2170f0c-Abstract.html}
  {A simple language model for task-oriented dialogue}.
\newblock In \emph{Advances in Neural Information Processing Systems 33: Annual
  Conference on Neural Information Processing Systems 2020, NeurIPS 2020,
  December 6-12, 2020, virtual}.

\bibitem[{Kingma and Ba(2015)}]{kingma2014adam}
Diederik~P. Kingma and Jimmy Ba. 2015.
\newblock \href {http://arxiv.org/abs/1412.6980} {Adam: {A} method for
  stochastic optimization}.
\newblock In \emph{3rd International Conference on Learning Representations,
  {ICLR} 2015, San Diego, CA, USA, May 7-9, 2015, Conference Track
  Proceedings}.

\bibitem[{Larson et~al.(2019)Larson, Mahendran, Peper, Clarke, Lee, Hill,
  Kummerfeld, Leach, Laurenzano, Tang, and Mars}]{larson2019evaluation}
Stefan Larson, Anish Mahendran, Joseph~J. Peper, Christopher Clarke, Andrew
  Lee, Parker Hill, Jonathan~K. Kummerfeld, Kevin Leach, Michael~A. Laurenzano,
  Lingjia Tang, and Jason Mars. 2019.
\newblock \href {https://doi.org/10.18653/v1/D19-1131} {An evaluation dataset
  for intent classification and out-of-scope prediction}.
\newblock In \emph{Proceedings of the 2019 Conference on Empirical Methods in
  Natural Language Processing and the 9th International Joint Conference on
  Natural Language Processing (EMNLP-IJCNLP)}, pages 1311--1316, Hong Kong,
  China. Association for Computational Linguistics.

\bibitem[{Liu et~al.(2021)Liu, Eshghi, Swietojanski, and
  Rieser}]{liu2021benchmarking}
Xingkun Liu, Arash Eshghi, Pawel Swietojanski, and Verena Rieser. 2021.
\newblock Benchmarking natural language understanding services for building
  conversational agents.
\newblock In \emph{Increasing Naturalness and Flexibility in Spoken Dialogue
  Interaction}, pages 165--183. Springer.

\bibitem[{Liu et~al.(2019)Liu, Ott, Goyal, Du, Joshi, Chen, Levy, Lewis,
  Zettlemoyer, and Stoyanov}]{liu2019roberta}
Yinhan Liu, Myle Ott, Naman Goyal, Jingfei Du, Mandar Joshi, Danqi Chen, Omer
  Levy, Mike Lewis, Luke Zettlemoyer, and Veselin Stoyanov. 2019.
\newblock \href {https://arxiv.org/abs/1907.11692} {Roberta: A robustly
  optimized bert pretraining approach}.
\newblock \emph{ArXiv preprint}, abs/1907.11692.

\bibitem[{Mehri et~al.(2020)Mehri, Eric, and Hakkani-Tur}]{mehri2020dialoglue}
Shikib Mehri, Mihail Eric, and Dilek Hakkani-Tur. 2020.
\newblock \href {https://arxiv.org/abs/2009.13570} {Dialoglue: A natural
  language understanding benchmark for task-oriented dialogue}.
\newblock \emph{ArXiv preprint}, abs/2009.13570.

\bibitem[{Nguyen et~al.(2020)Nguyen, Zhang, Xia, and Yu}]{nguyen2020dynamic}
Hoang Nguyen, Chenwei Zhang, Congying Xia, and Philip Yu. 2020.
\newblock \href {https://doi.org/10.18653/v1/2020.findings-emnlp.108} {Dynamic
  semantic matching and aggregation network for few-shot intent detection}.
\newblock In \emph{Findings of the Association for Computational Linguistics:
  EMNLP 2020}, pages 1209--1218, Online. Association for Computational
  Linguistics.

\bibitem[{Peng et~al.(2020)Peng, Li, Li, Shayandeh, Liden, and
  Gao}]{peng2020soloist}
Baolin Peng, Chunyuan Li, Jinchao Li, Shahin Shayandeh, Lars Liden, and
  Jianfeng Gao. 2020.
\newblock \href {https://arxiv.org/abs/2005.05298} {Soloist: Few-shot
  task-oriented dialog with a single pretrained auto-regressive model}.
\newblock \emph{ArXiv preprint}, abs/2005.05298.

\bibitem[{Peters et~al.(2018)Peters, Neumann, Iyyer, Gardner, Clark, Lee, and
  Zettlemoyer}]{peters-etal-2018-deep}
Matthew~E. Peters, Mark Neumann, Mohit Iyyer, Matt Gardner, Christopher Clark,
  Kenton Lee, and Luke Zettlemoyer. 2018.
\newblock \href {https://doi.org/10.18653/v1/N18-1202} {Deep contextualized
  word representations}.
\newblock In \emph{Proceedings of the 2018 Conference of the North {A}merican
  Chapter of the Association for Computational Linguistics: Human Language
  Technologies, Volume 1 (Long Papers)}, pages 2227--2237, New Orleans,
  Louisiana. Association for Computational Linguistics.

\bibitem[{Radford et~al.(2019)Radford, Wu, Child, Luan, Amodei, and
  Sutskever}]{radford2019language}
Alec Radford, Jeffrey Wu, Rewon Child, David Luan, Dario Amodei, and Ilya
  Sutskever. 2019.
\newblock Language models are unsupervised multitask learners.
\newblock \emph{OpenAI blog}, 1(8):9.

\bibitem[{Wang et~al.(2019)Wang, Pruksachatkun, Nangia, Singh, Michael, Hill,
  Levy, and Bowman}]{DBLP:conf/nips/WangPNSMHLB19}
Alex Wang, Yada Pruksachatkun, Nikita Nangia, Amanpreet Singh, Julian Michael,
  Felix Hill, Omer Levy, and Samuel~R. Bowman. 2019.
\newblock \href
  {https://proceedings.neurips.cc/paper/2019/hash/4496bf24afe7fab6f046bf4923da8de6-Abstract.html}
  {Superglue: {A} stickier benchmark for general-purpose language understanding
  systems}.
\newblock In \emph{Advances in Neural Information Processing Systems 32: Annual
  Conference on Neural Information Processing Systems 2019, NeurIPS 2019,
  December 8-14, 2019, Vancouver, BC, Canada}, pages 3261--3275.

\bibitem[{Wei et~al.(2018)Wei, Liu, Peng, Tou, Chen, Huang, Wong, and
  Dai}]{wei-etal-2018-task}
Zhongyu Wei, Qianlong Liu, Baolin Peng, Huaixiao Tou, Ting Chen, Xuanjing
  Huang, Kam-fai Wong, and Xiangying Dai. 2018.
\newblock \href {https://doi.org/10.18653/v1/P18-2033} {Task-oriented dialogue
  system for automatic diagnosis}.
\newblock In \emph{Proceedings of the 56th Annual Meeting of the Association
  for Computational Linguistics (Volume 2: Short Papers)}, pages 201--207,
  Melbourne, Australia. Association for Computational Linguistics.

\bibitem[{Wu et~al.(2020)Wu, Hoi, Socher, and Xiong}]{wu2020tod}
Chien-Sheng Wu, Steven~C.H. Hoi, Richard Socher, and Caiming Xiong. 2020.
\newblock \href {https://doi.org/10.18653/v1/2020.emnlp-main.66} {{TOD}-{BERT}:
  Pre-trained natural language understanding for task-oriented dialogue}.
\newblock In \emph{Proceedings of the 2020 Conference on Empirical Methods in
  Natural Language Processing (EMNLP)}, pages 917--929, Online. Association for
  Computational Linguistics.

\bibitem[{Xia et~al.(2020{\natexlab{a}})Xia, Xiong, Yu, and
  Socher}]{xia2020composed}
Congying Xia, Caiming Xiong, Philip Yu, and Richard Socher. 2020{\natexlab{a}}.
\newblock \href {https://doi.org/10.18653/v1/2020.findings-emnlp.303} {Composed
  variational natural language generation for few-shot intents}.
\newblock In \emph{Findings of the Association for Computational Linguistics:
  EMNLP 2020}, pages 3379--3388, Online. Association for Computational
  Linguistics.

\bibitem[{Xia et~al.(2020{\natexlab{b}})Xia, Xiong, Yu, and
  Socher}]{xia-etal-2020-composed}
Congying Xia, Caiming Xiong, Philip Yu, and Richard Socher. 2020{\natexlab{b}}.
\newblock \href {https://doi.org/10.18653/v1/2020.findings-emnlp.303} {Composed
  variational natural language generation for few-shot intents}.
\newblock In \emph{Findings of the Association for Computational Linguistics:
  EMNLP 2020}, pages 3379--3388, Online. Association for Computational
  Linguistics.

\bibitem[{Yan et~al.(2017)Yan, Duan, Chen, Zhou, Zhou, and
  Li}]{yan2017building}
Zhao Yan, Nan Duan, Peng Chen, Ming Zhou, Jianshe Zhou, and Zhoujun Li. 2017.
\newblock \href {http://aaai.org/ocs/index.php/AAAI/AAAI17/paper/view/14261}
  {Building task-oriented dialogue systems for online shopping}.
\newblock In \emph{Proceedings of the Thirty-First {AAAI} Conference on
  Artificial Intelligence, February 4-9, 2017, San Francisco, California,
  {USA}}, pages 4618--4626. {AAAI} Press.

\bibitem[{Zhang et~al.(2020{\natexlab{a}})Zhang, Hashimoto, Liu, Wu, Wan, Yu,
  Socher, and Xiong}]{zhang2020discriminative}
Jianguo Zhang, Kazuma Hashimoto, Wenhao Liu, Chien-Sheng Wu, Yao Wan, Philip
  Yu, Richard Socher, and Caiming Xiong. 2020{\natexlab{a}}.
\newblock \href {https://doi.org/10.18653/v1/2020.emnlp-main.411}
  {Discriminative nearest neighbor few-shot intent detection by transferring
  natural language inference}.
\newblock In \emph{Proceedings of the 2020 Conference on Empirical Methods in
  Natural Language Processing (EMNLP)}, pages 5064--5082, Online. Association
  for Computational Linguistics.

\bibitem[{Zhang et~al.(2020{\natexlab{b}})Zhang, Lyu, and
  Callison-Burch}]{zhang2020intent}
Li~Zhang, Qing Lyu, and Chris Callison-Burch. 2020{\natexlab{b}}.
\newblock \href {https://aclanthology.org/2020.aacl-main.35} {Intent detection
  with {W}iki{H}ow}.
\newblock In \emph{Proceedings of the 1st Conference of the Asia-Pacific
  Chapter of the Association for Computational Linguistics and the 10th
  International Joint Conference on Natural Language Processing}, pages
  328--333, Suzhou, China. Association for Computational Linguistics.

\bibitem[{Zhang et~al.(2020{\natexlab{c}})Zhang, Takanobu, Zhu, Huang, and
  Zhu}]{zhang2020recent}
Zheng Zhang, Ryuichi Takanobu, Qi~Zhu, Minlie Huang, and Xiaoyan Zhu.
  2020{\natexlab{c}}.
\newblock Recent advances and challenges in task-oriented dialog systems.
\newblock \emph{Science China Technological Sciences}, pages 1--17.

\end{thebibliography}

\end{document}